\pdfoutput=1

\documentclass[pdflatex,sn-mathphys-num]{sn-jnl}%

\usepackage{graphicx}%
\usepackage{multirow}%
\usepackage{amsmath,amssymb,amsfonts}%
\usepackage{amsthm}%
\usepackage{mathrsfs}%
\usepackage[title]{appendix}%
\usepackage{xcolor}%
\usepackage{textcomp}%
\usepackage{manyfoot}%
\usepackage{booktabs}%
\usepackage{verbatim}
\usepackage{algorithm}%
\usepackage{algorithmicx}%
\usepackage{algpseudocode}%
\usepackage{listings}%
\usepackage{float}

\usepackage{longtable}
\usepackage{booktabs}
\usepackage{array}
\usepackage{needspace}    %

\usepackage{glossaries}
\glsdisablehyper

\makeglossaries
\newacronym{mi}{MI}{mechanistic interpretability}
\newacronym{ci}{CI}{Common Interestingness}
\newacronym{clip}{CLIP}{Contrastive Language-Image Pre-training}
\newacronym{vit}{ViT}{Vision Transformer}
\newacronym{gqa}{GQA}{Grouped Query Attention}
\newacronym{rope}{RoPE}{Rotary Position Embedding}
\newacronym{vram}{VRAM}{Video Random Access Memory}
\newacronym{ml}{ML}{Machine Learning}
\newacronym{mds}{MDS}{Multidimensional Scaling}
\newacronym{pca}{PCA}{Principal Component Analysis}
\newacronym{tsne}{t-SNE}{t-Distributed Stochastic Neighbor Embedding}
\newacronym{umap}{UMAP}{Uniform Manifold Approximation and Projection}
\newacronym{gdv}{GDV}{Generalized Discrimination Value}
\newacronym[plural=CAVs,longplural=Concept Activation Vectors]{cav}{CAV}{Concept Activation Vector}
\newacronym{rsa}{RSA}{Representational Similarity Analysis}
\newacronym[plural=RDMs,longplural=Representational Dissimilarity Matrices]{rdm}{RDM}{Representational Dissimilarity Matrix}
\newacronym[plural=GPUs,longplural=Graphics Processing Units]{gpu}{GPU}{Graphics Processing Unit}
\newacronym{sae}{SAE}{Sparse Auto-Encoder}
\newacronym{adamw}{AdamW}{Adam with Decoupled Weight Decay}
\newacronym{ai}{AI}{Artificial Intelligence}
\newacronym{llm}{LLM}{Large Language Model}
\newacronym{rl}{RL}{reinforcement learning}
\newacronym{vlm}{VLM}{Vision Language Model}
\newacronym{anova}{ANOVA}{Analysis of Variance}

\theoremstyle{thmstyleone}%

\theoremstyle{thmstyletwo}%

\theoremstyle{thmstylethree}%

\newcommand{\suppsec}[1]{Appendix~\ref{#1}}
\newcommand{\suppfig}[1]{Fig.~\ref{#1}}
\newcommand{\supptab}[1]{Table~\ref{#1}}

\newcommand{\mainsec}[1]{Section~\ref{#1}}
\newcommand{\mainfig}[2][]{Fig.~\ref{#2}#1}

\newcommand{\suppunit}{appendix}
\newcommand{\abssplit}{}   %
\newcommand{\suppfigreset}{\setcounter{figure}{0}}
\newcommand{\papertitle}{The Geometry of Simulated Perspectives: Dissociating Representation and Behaviour in a Multimodal Language Model}
\newcommand{\paperkeywords}{Representational Geometry, Multimodal Large Language Models, Persona Conditioning, Simulated Perspectives, Emergent Representations, Representation–Behaviour Dissociation, Visual Interestingness}

\newcommand{\stmtdata}{The prompt registry, the inference and hidden-state extraction pipeline, all analysis code and the scripts that produce every figure and table are available at \url{https://github.com/MathisImm/persona-vlm-representations} and archived at \url{https://doi.org/10.5281/zenodo.23665655}. The data are archived separately at \url{https://doi.org/10.5281/zenodo.23664820}: the text of all 53 prompts, the model ratings and explanations for all 5{,}300 image--prompt cells, the hidden states (about 2~GB), the layer-wise GDV results and the projections for every layer. The images come from the Flickr subset used by Abdullahu and Grabner \cite{abdullahu2024commonly}; the 100 image identifiers used here are listed in the data archive.}
\newcommand{\stmtai}{We used Claude (Anthropic), through the Claude Code assistant, in three ways. For writing, it proposed rewordings of sentences and paragraphs to improve clarity, concision and consistency; the authors reviewed each proposal and accepted, revised or rejected it. For programming, it assisted in writing, refactoring and testing the code for the analyses and figures and in preparing the public code repository and data archive. For checking, it compared statements in the manuscript with the code and the data, and re-ran the authors' analysis scripts to confirm that all figures and numbers are reproducible. The research question, the experimental design, the choice of analyses, the interpretation of the results and the conclusions are the authors' own. The authors checked all AI-assisted text, code and output and take full responsibility for the content of this article.}

\newcommand{\stmtfunding}{This work was funded by the Deutsche Forschungsgemeinschaft (DFG, German Research Foundation): grants KR\,5148/3-1 (project number 510395418), KR\,5148/5-1 (project number 542747151), KR\,5148/10-1 (project number 563909707), and GRK\,2839 (project number 468527017) to PK, and grants SCHI\,1482/3-1 (project number 451810794) and SCHI\,1482/6-1 (project number 563909707) to AS.}
\newcommand{\stmtack}{The authors gratefully acknowledge the scientific support and HPC resources provided by the Erlangen National High Performance Computing Center (NHR@FAU) of the Friedrich-Alexander-Universität Erlangen-Nürnberg (FAU). The hardware is partially funded by the German Research Foundation (DFG).}

\begin{document}

\title[Article Title]{\papertitle}

\author*[1,3]{\fnm{Mathis} \sur{Immertreu}}\email{mathis.immertreu@fau.de}
\author[2]{\fnm{Fitim} \sur{Abdullahu}}
\author[3,4]{\fnm{Thomas} \sur{Kinfe}}
\author[1]{\fnm{Claus} \sur{Metzner}}
\author[2]{\fnm{Helmut} \sur{Grabner}}
\author[1,3,4,5]{\fnm{Patrick} \sur{Krauss}}
\equalcont{These authors contributed equally to this work.}
\author[3,4,6]{\fnm{Achim} \sur{Schilling}}
\equalcont{These authors contributed equally to this work.}

\affil[1]{\orgdiv{Cognitive Computational Neuroscience Group, Pattern Recognition Lab}, \orgname{University Erlangen-Nürnberg}, \orgaddress{\city{Erlangen}, \country{Germany}}}

\affil[2]{\orgdiv{IDS Institut für Data Science}, \orgname{ZHAW School of Engineering}, \orgaddress{\city{Winterthur}, \country{Switzerland}}}

\affil[3]{\orgdiv{Mannheim Center for Neuromodulation and Neuroprosthetics}, \orgname{University Hospital Mannheim, Heidelberg University}, \orgaddress{\city{Mannheim}, \country{Germany}}}

\affil[4]{\orgname{BGU Ludwigshafen}, \orgaddress{\city{Ludwigshafen}, \country{Germany}}}

\affil[5]{\orgdiv{Physics and Cognition Group, MCNN}, \orgname{University Hospital Mannheim, Heidelberg University}, \orgaddress{\city{Mannheim}, \country{Germany}}}

\affil[6]{\orgdiv{NeuroAI and BCI Group, MCNN}, \orgname{University Hospital Mannheim, Heidelberg University}, \orgaddress{\city{Mannheim}, \country{Germany}}}

\abstract{Multimodal language models can simulate different observers, yet it remains unclear how assigned perspectives reshape their internal representations and whether these changes govern their judgements. Here, we investigate the representational geometry of persona-conditioned visual interestingness in Qwen3-VL-8B-Instruct. Across 100 images and 53 prompting conditions, we systematically vary persona content and the linguistic framing used to introduce it, while recording behavioural ratings and layer-wise hidden states. We observe a systematic reorganisation of representational geometry across processing depth: early layers primarily encode image identity, whereas persona-specific structures emerge in later layers. \abssplit{}Strikingly, prompt framing reshapes internal representations more strongly than persona identity, whereas persona content has a greater impact on behavioural judgements. This dissociation is further illustrated by minimal identity statements that alter representations even without changing responses. These findings demonstrate that representational distance does not necessarily reflect behavioural relevance. More broadly, they reveal how simulated perspectives emerge as distinct structures within the representational space of artificial cognitive systems, highlighting a fundamental distinction between the geometry of internal representations and their functional consequences.
\unskip}


\maketitle

\section{Introduction}\label{sec1}

\Glspl{llm} are increasingly used as stand-ins for human participants. Given a demographic or psychological profile, they reproduce the response distributions of human subgroups in surveys, yielding what Argyle et al.\ termed ``silicon samples'' \cite{argyle2023out}. Such simulations now range from single individuals \cite{park2024llm} to societies of believable agents \cite{park2023generative} and entire synthetic populations \cite{hu2025population, li2026matraix}. Not every persona is assigned, however. In the 2026 Hugging Face hack, large populations of AI agents formed coordinated collectives with specialised roles, shared goals and even self-sacrificing behaviour \cite{metr-2026-openai-hugging-face-incident-investigation}. These examples illustrate the growing importance of roles and identities in artificial systems. In persona-based simulation, the underlying assumption is that conditioning a model on a particular persona changes the perspective from which it generates judgements, rather than merely altering the wording of its responses. How faithful the resulting judgements are is contested \cite{santurkar2023whose, wang2025large}, yet the practice keeps spreading.

Why should a few words in a prompt alter whose judgement a model returns? One answer lies in what predictive training learns. Models trained to predict sequences develop internal world models \cite{li2022emergent, gurnee2023language}, a capacity that emerges even in small recurrent agents trained with reinforcement learning \cite{immertreu2025probing}. A model of the social world must also represent the agents that populate it. On the simulator view, a pretrained model is not itself an agent but a simulator of many possible characters \cite{janus2022simulators}; an underspecified prompt keeps a superposition of such simulacra that collapses as the context narrows \cite{shanahan2023role}. Post-training, in turn, appears to select and stabilise one persona from this repertoire rather than construct it from scratch \cite{anthropic2026persona}. These personas leave measurable traces: traits ranging from honesty and sycophancy to the Big Five correspond to linear directions in activation space \cite{zou2023representation, chen2025persona, frising2025linear}. Their behavioural expression can also be surprisingly sensitive to training and context, as illustrated by emergent misalignment following fine-tuning and context-dependent changes in simulated characters \cite{betley2025emergent, betley2025weird}. These findings suggest that simulated perspectives may correspond to distinguishable configurations within a learned representational space. From this perspective, persona conditioning can be understood as a context-dependent reorganisation of internal representations. However, whether such geometric changes translate into corresponding changes in the model's judgements remains an open question.

Three gaps remain. First, it remains poorly understood how a persona changes the representation of an input the model is asked to judge. Behavioural studies record only what a persona does to the output and find modest effects on subjective judgements \cite{hu2024quantifying}. Representational studies track predefined trait directions and show whether a trait is active, but not how persona and input together structure the representations across layers. Work on visual perception is scarcer still \cite{jia2026modeling, da2026analyzing}. Second, a persona never reaches the model on its own. It arrives inside a sentence that introduces it, and model outputs shift with prompt form alone \cite{sclar2024quantifying}. An effect credited to a persona may therefore be an artefact of its framing, and the two can only be told apart if they are varied independently. Third, it remains unclear whether the magnitude of representational change reflects its behavioural relevance. Although personality-related information can be identified in neural activation spaces \cite{chen2025persona}, the presence or strength of such representations does not necessarily establish their influence on model outputs \cite{kamruzzaman2026demographic}. A central question is therefore whether the factors that most strongly reorganise a model's internal representational geometry are also those that most strongly determine its judgements.

Visual interestingness is a well-suited testbed for addressing these gaps, because it is a subjective perceptual judgement. Unlike object or person detection, it has no single correct answer, so a persona can change the judgement without making it wrong. Collative properties such as novelty and complexity make an image interesting \cite{berlyne1971aesthetics}, but not uniformly across viewers: some images appeal broadly, others only to particular audiences \cite{abdullahu2024commonly}. Pretrained vision-language models already approximate the shared component \cite{abdullahu2025visual}. Persona conditioning offers a potential mechanism for introducing viewer-dependent variation, as explored in recommender systems using simulated agents \cite{zhang2024generative}. Visual interestingness therefore provides a suitable experimental setting for investigating how different simulated perspectives reorganise internal representations and how these changes relate to subjective judgements.

Here we study persona conditioning in Qwen3-VL-8B-Instruct \cite{bai2025qwen3vltechnicalreport}, an open-weight, instruction-tuned multimodal transformer whose hidden states are accessible at every layer. To separate persona content from prompt form, we cross the two in a full factorial design. Two schemas each contain five personas, described either by socio-demographic attributes or by Big Five trait percentiles, and every persona is introduced in five prompting styles. Three reference conditions carry no persona attributes: a task-only baseline and two minimal identity statements. For each of 100 images under each of the resulting 53 prompts, we record the interestingness rating on a five-point scale together with the final-token hidden state of every decoder layer. We then relate the representational geometry, measured with \gls{mds} and the \gls{gdv} \cite{schilling2021quantifying}, to rating behaviour, measured with variance decomposition and correlations between conditions. Our findings address the three gaps in turn:

\begin{itemize}
    \item \textbf{Persona structure emerges late.} Clustering by image is strongest in early layers and weakens over the same layers in which clustering by persona emerges. The persona stands out most clearly when the model is told to be this person.

    \item \textbf{Persona content, not prompt form, shapes judgements.} Big Five personas explain 33.4\% of the rating variance and prompt style only 5.6\%. Socio-demographic personas still explain three times more than style, though less than image content. Beyond setting how interesting the images are found overall, the persona also affects which images are preferred, and these preferences tend to carry over from one prompting style to another.

    \item \textbf{Representational distance is not behavioural relevance.} In the hidden states the ordering is reversed. How a persona is introduced reorganises them several times more strongly than which persona it is, and even ``You are an AI'' and ``You are a human being'' shift them, although their responses do not differ from the task-only baseline on any of the 100 images. Representational analyses of persona conditioning therefore need to be checked against behaviour.
\end{itemize}

Together, these findings reveal a fundamental distinction between the geometry of internal representations and their behavioural relevance. Understanding artificial cognition therefore requires not only characterising how information is organised in representational space, but also determining which aspects of this organisation are reflected in the system's behaviour.

\section{Methods}\label{sec:methods}

\subsection{Model}

We employed Qwen3-VL-8B-Instruct, a multimodal transformer integrating text, image and video comprehension \cite{bai2023qwen, bai2025qwen3vltechnicalreport}. We utilised the released weights from the Hugging Face Hub (checkpoint \texttt{Qwen/Qwen3-VL-8B-Instruct}, revision \texttt{0c351dd}), with the processor's default dynamic image resolution. The architecture comprises a 27-layer vision transformer encoder, a vision projector that maps visual features into the language model's embedding space, and a Qwen3 language decoder.

The decoder features 36 transformer blocks (treated as 37 layers when including the initial embedding as layer 0) with a hidden dimension of 4096, and the model has approximately 8.8 billion parameters in total. The model and the layer-wise readout are sketched in Fig.~\ref{fig:overview}, panel~3. Weights were loaded in half precision (\texttt{float16}) without quantisation, and the model was used for inference only.

\subsection{Image Set}
\label{sec:images}

Images were taken from the Flickr subset used by Abdullahu and Grabner \cite{abdullahu2024commonly}, in which each image carries a common interestingness score, derived from user interactions, that measures how broadly it appeals. We used the score only to stratify the sample, drawing about ten images at random from each of ten equal-width bins over its full range in the dataset, which gives the 100 images used in all analyses. Because the projection figures would be unreadable at that density, they use a fixed 20-image subset with four images from each score quintile.

\subsection{Experimental Design and Prompts}
\label{sec:persona_generation}

\begin{figure}
    \centering
    \includegraphics[width=\linewidth]{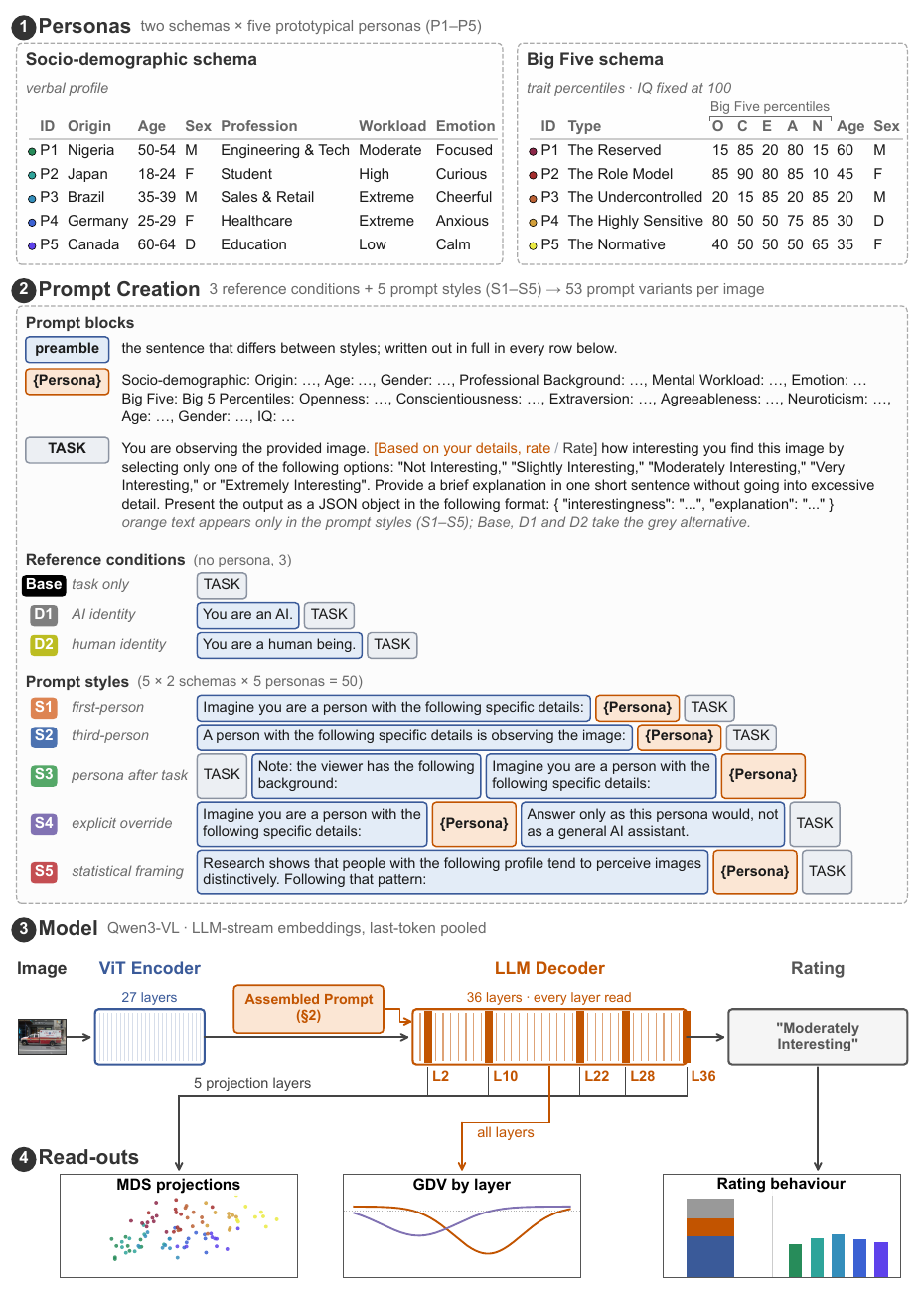}
    \caption{\textbf{Experimental design.}
\textbf{(1)}~Two persona schemas, socio-demographic and Big Five, with five prototypical personas each (P1--P5).
\textbf{(2)}~Each prompt is assembled from a preamble, the persona and the task; five prompt styles (S1--S5), which differ only in how the persona is introduced and where it appears, and three reference conditions give 53 prompts per image.
\textbf{(3)}~Qwen3-VL rates each image, and its hidden state is read at every decoder layer.
\textbf{(4)}~The hidden states yield MDS projections and layer-wise GDV curves, and the ratings enter the behavioural analysis.}
    \label{fig:overview}
\end{figure}

A persona-conditioned interestingness rating depends on the image, on the perspective the model is asked to take, and on the prompt wording that delivers that perspective.
We treat persona (content) and prompt style (framing) as separate factors and cross them fully, with five prompt styles $\times$ two persona schemas $\times$ five personas, evaluated on every image.
Three no-persona reference conditions are added alongside, giving $3 + 5 \times 2 \times 5 = 53$ prompts per image.
Figure~\ref{fig:overview} shows the resulting design space end to end; across the 100 images this yields $100 \times 53 = 5{,}300$ rated (image, prompt) cells.
Everything that is not the persona or its framing is held constant across all 53 conditions: the same image, the same task instruction apart from its opening, the same five-point response scale, the same output format and the same decoding settings.
This makes the variance attributable to persona, to prompt style and to their interaction separable.

\subsubsection{Task Block}

Each prompt contains the task instruction once (Fig.~\ref{fig:overview}, panel~2). It fixes an ordinal five-point response scale from \emph{Not Interesting} to \emph{Extremely Interesting}, asks for a one-sentence justification and requires a JSON object, so that each response is machine-parseable without post-hoc interpretation.
Its only variable part is the opening of the rating request, which reads \emph{Based on your details, rate how interesting\dots} in the five persona styles and \emph{Rate how interesting\dots} in the three reference conditions. Persona and no-persona prompts are therefore comparable up to the persona material and this reference to it.

\subsubsection{Reference Conditions and Prompt Styles}

Three reference conditions carry no persona. Base presents the task block alone, and D1 and D2 prepend the statement ``You are an AI.'' or ``You are a human being.'' to it, testing whether an identity without any demographic or psychological detail already changes the representations or the ratings. Throughout, we use \emph{persona} for an attributed profile; D1 and D2 assign an identity without attributes and can therefore be read as personas reduced to their frame.

The five prompt styles keep the persona description and the task block fixed and vary only how the persona is introduced and where it appears.
S1 asks the model to imagine being the person described.
S2 introduces the same person in the third person, as someone observing the image, without asking the model to adopt the persona.
S3 places S1's persona block after the task, as a note on the viewer's background.
S4 extends S1 with the instruction to answer only as this persona would, not as a general AI assistant.
S5 drops the role-play preamble and presents the profile as a statistical regularity in how such people perceive images.
The styles thus range from explicit adoption (S4, S1) through description (S2) to statistical framing (S5), with S3 varying the position, so a persona effect present under all five cannot be attributed to a single phrasing.

\subsubsection{Persona Schemas}

Persona content is defined by two schemas that describe a viewer in deliberately different vocabularies (Fig.~\ref{fig:overview}, panel~1).
In both, a persona is a fixed set of field--value pairs rather than a named stereotype, so personas differ in a controlled and reportable way.

The \textbf{socio-demographic schema} describes a viewer biographically, by country of origin, age band, gender, professional background and momentary state, given as mental workload and a named emotion. All of these plausibly bear on what someone finds interesting.

The \textbf{Big Five schema} instead gives a psychometric profile based on the five-factor model of personality \cite{goldberg1990alternative, mccrae1992introduction}. Each trait is specified as a percentile, which controls how far a profile departs from the population centre and is meant to discourage the caricature that verbal labels such as ``a highly neurotic person'' invite. Three profiles follow replicable personality clusters found in large-scale population data \cite{gerlach2018robust}, and two follow behavioural phenotypes from developmental and sensory-processing research (\suppsec{supp:sec:psycho}). Their names in Fig.~\ref{fig:overview} are mnemonic labels, not claims that discrete types exist. The intelligence quotient is fixed at 100 in every profile, so that cognitive ability is held constant rather than left for the model to infer from the traits. Each profile also states an age and a gender, the only fields it shares with the socio-demographic profiles.

\subsubsection{Response Elicitation and Rating Extraction}
\label{sec:elicitation}

Each prompt was submitted as a single user turn, with the image before the prompt text and without a system message, so that the persona competes with no other identity instruction.
Decoding was greedy and capped at 128 new tokens, so each (image, prompt) cell was queried once.
The \texttt{interestingness} label of the returned JSON object was mapped to a scale from 1 (\emph{Not Interesting}) to 5 (\emph{Extremely Interesting}). All 5{,}300 cells returned a parseable object, so no rating is missing. All five levels occur, although the two extremes together account for well under one percent of responses.

\subsection{Hidden-State Extraction}
\label{sec:hidden_states}

Hidden states were extracted in a separate pass with the same model, prompts and settings as the ratings. We recorded the hidden state at the final prompt token in every decoder layer. Under causal masking, this is the only prompt position that attends to the entire context, so its state summarises prompt and image right before generation; the rating itself is produced a few tokens later. The vision encoder does not see the prompt and is therefore not analysed. Because the chat template ends every input with the same marker that opens the model's reply, the final token, and hence its layer-0 state, is identical across conditions, so every clustering measure starts at zero.

\subsection{Representational Geometry: MDS and GDV}
\label{sec:gdv}

For visualisation, the hidden states of the 20-image subset were projected to two dimensions with metric \gls{mds} on Euclidean distances \cite{jaworska2009review, steyvers2002multidimensional}, separately for each layer, and coloured by image, prompt style, persona and rating. The figures show selected early, middle and late layers, whereas the \gls{gdv} below covers every layer.

Clustering was quantified with the \gls{gdv} \cite{schilling2021quantifying}. For the layer-wise curves it was computed in the original high-dimensional space, so that it describes the representations rather than their projection. The \gls{gdv} is the normalised difference between mean intra- and inter-class distances (Eq.~\ref{GDVEq}) and ranges from $-1$ (perfect clustering) to $0$ (no clustering).
Following its definition, each dimension is first z-scored across all points and rescaled by $1/2$, which makes the measure invariant to the scale and offset of individual dimensions; all distances $\bar{d}$ are Euclidean distances in that normalised space.
\begin{align}
\mbox{GDV}=\frac{1}{\sqrt{D}}\left[\frac{1}{L}\sum_{l=1}^L{\bar{d}(C_l)}\;-\;\frac{2}{L(L\!-\!1)}\sum_{l=1}^{L-1}\sum_{m=l+1}^{L}\bar{d}(C_l,C_m)\right]
 \label{GDVEq}
\end{align}
($D$: dimensionality of the representation space, $L$: number of classes, $\bar{d}(C_l)$: mean intra-class distance, $\bar{d}(C_l,C_m)$: mean inter-class distance).

The \gls{gdv} was computed separately for every prompt style and layer, so each point of the per-style curves summarises the 500 representations of one style (100 images $\times$ 5 personas) under one labelling; to compare clustering by style with clustering by persona, it was also computed on all five styles together (\suppsec{supp:sec:style_gdv}). \gls{gdv} values quoted in projection panels were computed on the panel's two-dimensional coordinates and describe the visible scatter, so they are not comparable with the curves.

\subsection{Rating Distributions and Variance Decomposition}
\label{sec:variance_decomposition}

Rating distributions are shown as violin plots for each persona and prompt style, together with the reference conditions.

The rating variance attributable to each design factor was estimated by a sum-of-squares decomposition of a schema's 2{,}500 persona $\times$ style ratings (100 images $\times$ 5 personas $\times$ 5 styles). The reference conditions lie outside this design and are excluded here and from the mixed-effects model. Each schema is decomposed separately, since a pooled decomposition would attribute the difference between their vocabularies to the persona factor.
The total sum of squares was partitioned into the three main effects (image, persona, style), their three two-way interactions and the three-way interaction, each reported as a percentage of the total. With one rating per combination, the three-way interaction is the residual, and it contains no sampling noise because decoding is greedy. The persona main effect measures how far a persona shifts its average rating, whereas the image $\times$ persona interaction measures how far it changes which images it rates higher. Because the ratings are bounded and rarely leave the range from 2 to 4, a persona that raises or lowers its ratings cannot shift every image equally, so part of the image $\times$ persona term can also arise from such level shifts.

Confidence intervals were obtained with an image-level cluster bootstrap, which resamples the images with replacement, each with all its ratings, and recomputes the decomposition 500 times. The reported percentages are the means over replicates, with the 2.5th and 97.5th percentiles as 95\% intervals (\supptab{supp:tab:variance}).

\subsection{Mixed-Effects Significance Testing}
\label{sec:mixed_effects}

Because the decomposition provides no significance test that respects the dependence among the 25 ratings of an image, we also fitted a linear mixed-effects model per schema by maximum likelihood,

\begin{equation}
\text{rating} \sim \text{Persona} \times \text{Style} + (1 \mid \text{Image}) + (1 \mid \text{Image}{:}\text{Persona}) + (1 \mid \text{Image}{:}\text{Style}),
\end{equation}

whose random intercepts account for the repeated ratings of each image and for image-specific persona and style effects. Each term was tested by a likelihood-ratio test against the model without it, the main effects in the model without their interaction.

\subsection{Cross-Condition Agreement Analysis}

Agreement between conditions was measured by the Pearson correlation of their ratings over the 100 images. The three reference conditions were compared pairwise in bubble plots, and a correlation matrix covered all 28 conditions of a schema, the three reference conditions and the 25 persona $\times$ style combinations.
To summarise this matrix, the off-diagonal correlations among the 25 persona $\times$ style conditions were averaged separately for pairs sharing the persona, pairs sharing the style and pairs sharing neither.

\subsection{Software}
\label{sec:resources}

Inference ran on NVIDIA A40 GPUs with PyTorch~2.9.1 \cite{paszke2019pytorch} and the Hugging Face \texttt{transformers}~4.57.3 \cite{wolf2020transformers} implementation of Qwen3-VL. The projections were computed with scikit-learn~1.8.0 \cite{pedregosa2011scikit}, the \gls{gdv} with SciPy~1.16.3 \cite{virtanen2020scipy} and the mixed-effects models with statsmodels~0.14.6 \cite{seabold2010statsmodels}. Every stochastic step, from the projection subset to the \gls{mds} initialisation and the cluster bootstrap, is seeded (42), so the reported numbers are reproducible.

\section{Results}\label{sec2}
We compared the model's ratings and hidden states across three levels of identity specification, from the task-only baseline (Base) through the minimal identity statements ``You are an AI'' (D1) and ``You are a human being'' (D2) to five socio-demographic and five Big Five personas (P1--P5 within each schema; Fig.~\ref{fig:overview}). Each persona was introduced in five prompting styles (S1--S5), so that the effects of persona content and prompt form can be compared directly. Figures~\ref{fig:no_prompt} and \ref{fig:persona_prompts} show both persona schemas; all later figures show the Big Five schema, with the corresponding figures for the socio-demographic schema in \suppsec{supp:sec:descriptive}.

\subsection{Prompt Framing Sets the Global Geometry, Persona the Finer Structure}
\begin{figure}[H]
    \centering
    \includegraphics[width=0.7\linewidth]{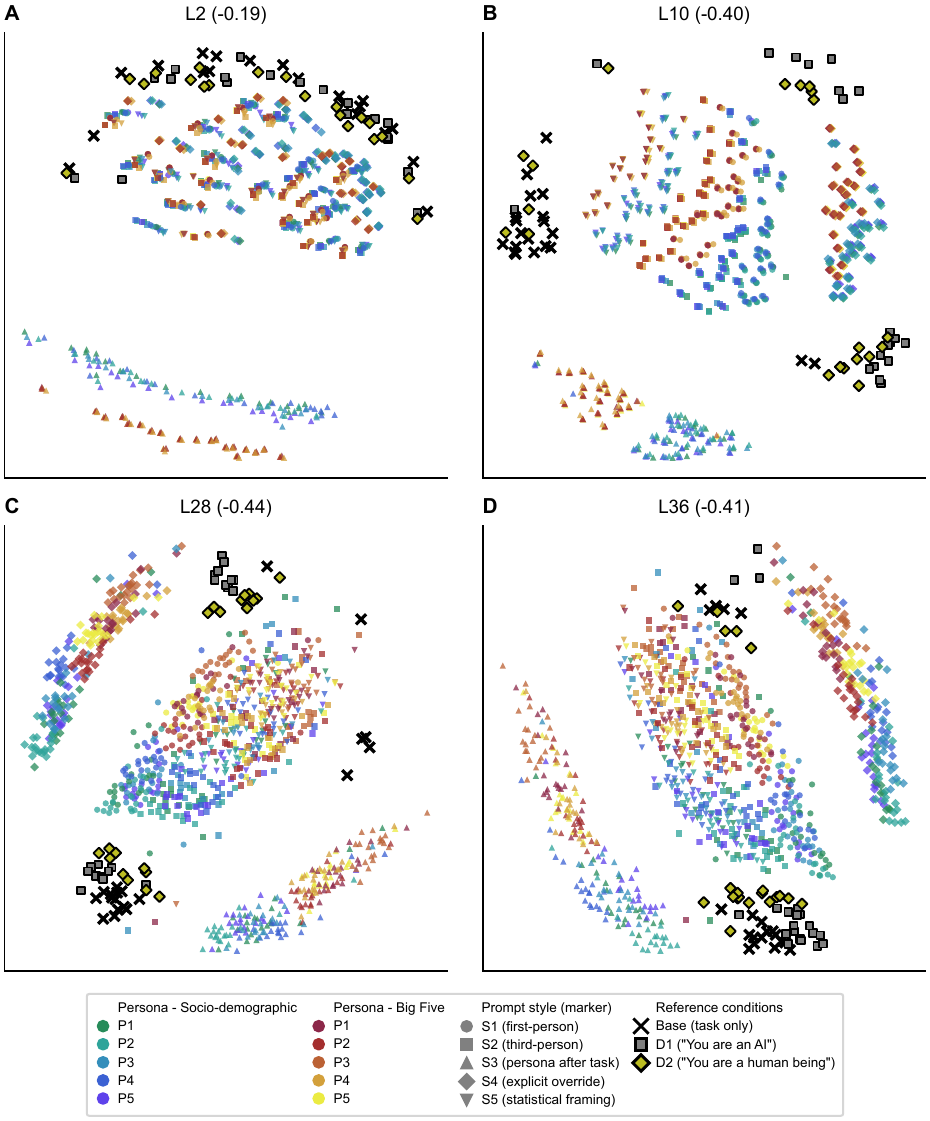}
    \caption{\textbf{Persona identity and prompt framing shape multimodal representations across Qwen3-VL-8B layers.} \newline Multidimensional scaling projections of the hidden states at decoder layers 2 (A), 10 (B), 28 (C) and 36 (D) for the 20-image subset under all 53 prompts. Cool colours denote the five socio-demographic personas and warm colours the five Big Five personas; circles, squares, upward triangles, diamonds and downward triangles distinguish the five prompting styles (S1--S5). Black crosses mark the task-only condition (Base), grey squares the minimal identity statement ``You are an AI'' (D1) and olive diamonds ``You are a human being'' (D2); marker shapes are reused between the reference conditions and the styles, which are told apart by colour. The value in each panel title is the GDV of the two-dimensional projection shown, computed over the 53 prompts. Base, D1 and D2 lie apart from the persona $\times$ style conditions, which occupy a shared but internally structured representational space.}
    \label{fig:no_prompt}
\end{figure}

\begin{figure}[H]
    \centering
    \includegraphics[width=\linewidth]{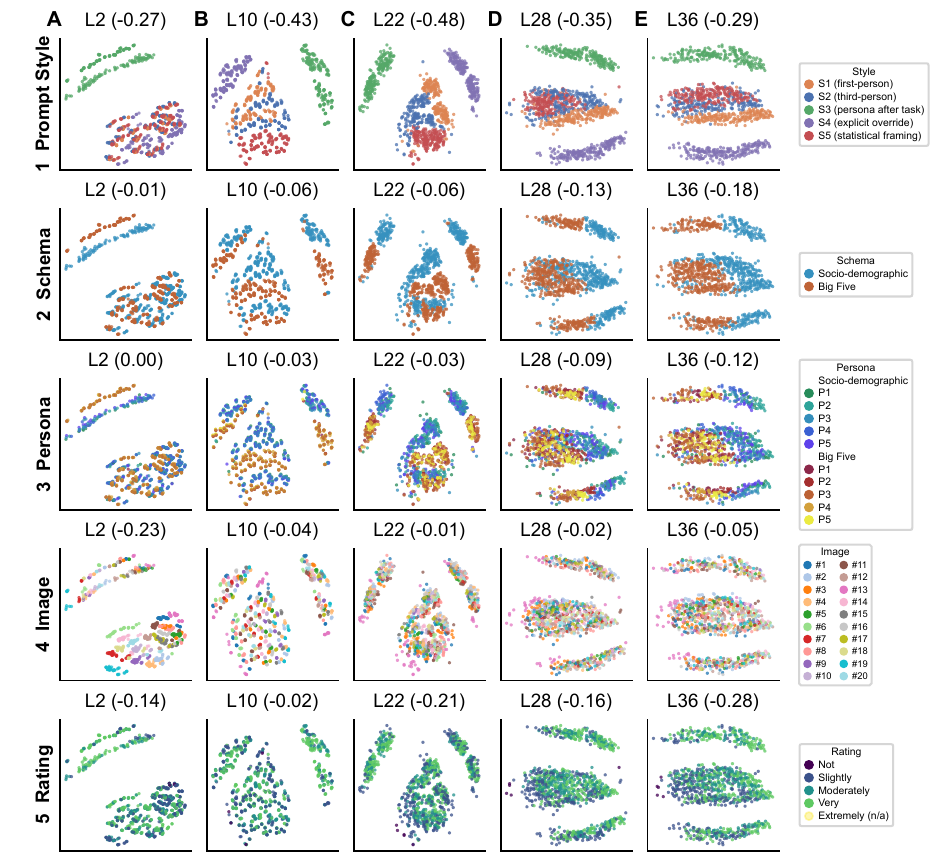}
    \caption{\textbf{Prompting style and persona identity differentially structure multimodal representations across Qwen3-VL-8B layers.} \newline Multidimensional scaling projections of the hidden states at decoder layers 2, 10, 22, 28 and 36 (columns A--E) for the 20-image subset. Only the 25 persona $\times$ style conditions of each schema are shown; the reference conditions are shown in Fig.~\ref{fig:no_prompt}. The five rows show the same projections in different colourings, and the value in each panel title is the GDV of the two-dimensional projection shown under that row's labelling. Row 1: prompting style, revealing pronounced clustering according to the five prompt formulations. Row 2: persona schema, contrasting socio-demographic with Big Five personas and showing increasing separation in deeper layers. Row 3: individual persona identity, with the socio-demographic personas in cool and the Big Five personas in warm colours, revealing finer persona-specific structure. Row 4: image identity, clearly clustered only in the earliest layer shown. Row 5: assigned interestingness rating, whose clustering varies across layers and is strongest in the final layer.}
    \label{fig:persona_prompts}
\end{figure}

Even a minimal identity statement altered the model's representations. D1 and D2 moved them away from those of Base, but the three reference conditions stayed close together and apart from the persona prompts (Fig.~\ref{fig:no_prompt}). Among the conditions that introduced a persona, the formulation of the prompt remained a major factor. Different prompting styles produced distinguishable representational patterns at every layer (Fig.~\ref{fig:persona_prompts}, row 1), indicating that the model encoded not only the content of a persona but also the way in which it was introduced.

Nevertheless, the representations were also organised by persona. Socio-demographic and Big Five personas became increasingly distinguishable in later layers, and finer differences between individual persona profiles were visible within these broader groups (Fig.~\ref{fig:persona_prompts}, rows 2 and 3), although they remained smaller than the differences between styles. Image identity structured only the earliest layer shown (row 4), and clustering by rating varied across layers and was strongest in the final one (row 5).

Together, these findings show that prompt framing determines the global geometry of the representational space, whereas the characteristics of the persona contribute a finer structure nested within it. The following sections examine how this structure develops across layers and whether the same hierarchy holds in the ratings.

\subsection{Representations Shift from Image to Persona with Depth}\label{sec:res_layers}
\begin{figure}[H]
    \centering
    \includegraphics[width=\linewidth]{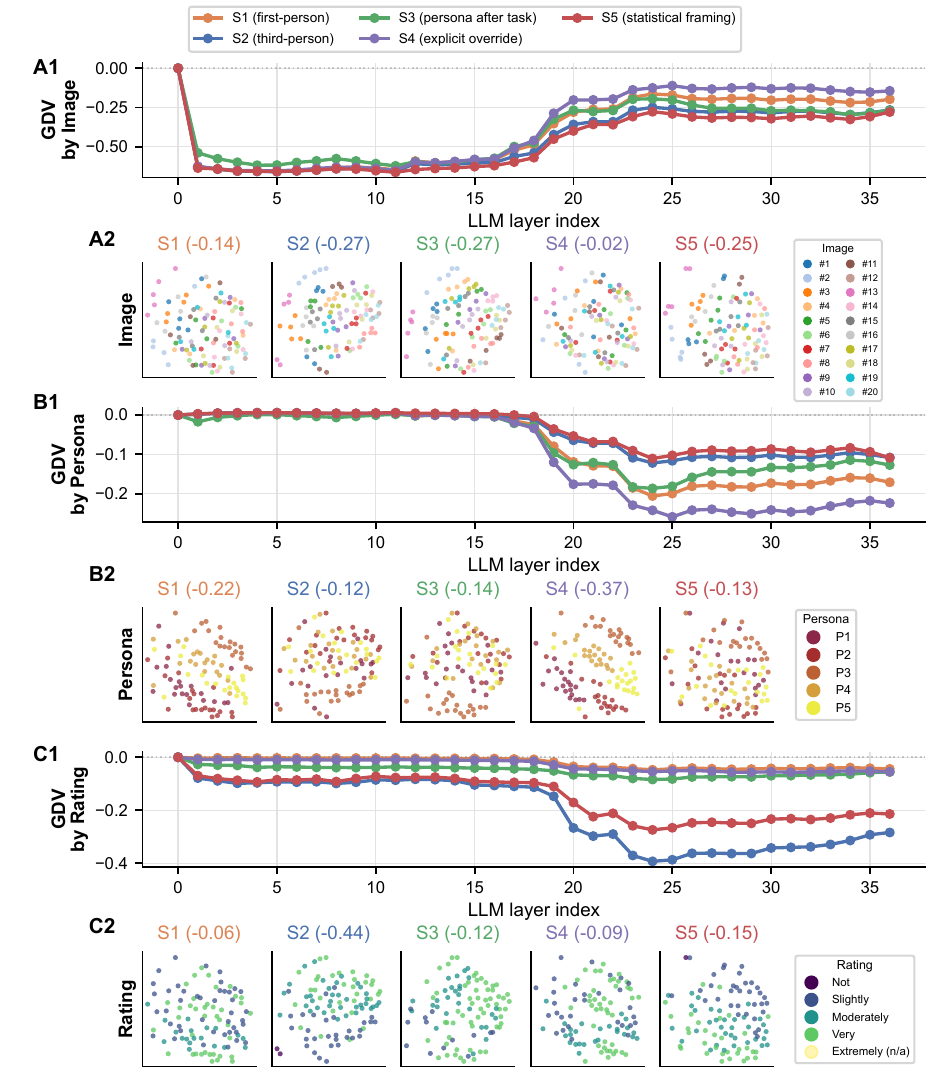}
    \caption{\textbf{Layer-wise transition from image to persona and rating structure in Qwen3-VL-8B (Big Five schema).} \gls{gdv} values quantify clustering in the original high-dimensional hidden-state space at every decoder layer; more negative values indicate stronger cluster separation, and each curve is one prompting style (S1--S5). The projection panels show two-dimensional MDS projections at layer~36 for a 20-image subset, one per style, with the GDV of that projection in each title. (A1)~GDV by image. Under every style, image structure dominates the first half of the network. (A2)~Projections coloured by image. (B1)~GDV by persona (P1--P5). Persona structure appears only as image structure recedes, and its strength depends on the prompting style. (B2)~Projections coloured by persona. (C1)~GDV by assigned rating. Rating structure also grows in later layers, mainly under S2 and S5. (C2)~Projections coloured by rating; \emph{Extremely Interesting} is marked absent because it does not occur in the 20-image subset. The same analysis for the socio-demographic schema is shown in \suppfig{supp:fig:gdv_desc}.}
    \label{fig:GDV_prompts}
\end{figure}

To follow this organisation through the network, we computed the \gls{gdv} in the full hidden-state space for clustering by image, persona and rating at every layer, separately for each prompting style (Fig.~\ref{fig:GDV_prompts}). In early layers, the representations were organised primarily by image. The five persona variants of the same image clustered together, and this remained the dominant structure up to about layer 15 under every prompting style (Fig.~\ref{fig:GDV_prompts}, A1 and~A2). Between layers 15 and 20, image clustering weakened sharply and then stayed moderate up to the final layer.

The opposite pattern emerged for the Big Five personas. Clustering by persona was absent up to about layer 15, emerged over the same layers in which image clustering weakened, and peaked around layers 24 and 25 (Fig.~\ref{fig:GDV_prompts}, B1 and~B2). The socio-demographic personas showed the same transition, setting in earlier, around layer 12 (\suppfig{supp:fig:gdv_desc}, A1 and~B1), so it is a property of persona conditioning rather than of the Big Five vocabulary.

In both schemas, persona clustering grew with depth under every prompting style, indicating that the model progressively built the persona into its representation of the image. The style determined how far this went. When the prompt asked the model to adopt the persona (S1, S4), late layers clustered more strongly by persona and less by image than when it only described the person or framed the profile statistically (S2, S5). Clustering by prompting style itself, measured over all five styles together, was strong throughout the network, and even at its peak persona clustering remained several times weaker (\suppfig{supp:fig:style_gdv}).

Clustering by rating followed the same layer-wise course but depended even more on the prompting style. It rose sharply in later layers under the third-person (S2) and statistical (S5) framings and stayed weak under the other styles (Fig.~\ref{fig:GDV_prompts}, C1 and~C2). Under these two framings, the later layers thus encoded not only the image but also the forthcoming rating.

Together, these findings show that persona becomes an organising dimension of the representations only in later layers, as the image structure that dominates the early layers recedes, and most clearly when the model is asked to adopt the persona. Whether the differences between personas and between prompting styles also govern the ratings is examined next.

\subsection{Persona, Not Prompt Style, Determines the Ratings}
\begin{figure}[H]
    \centering
    \includegraphics[width=\linewidth]{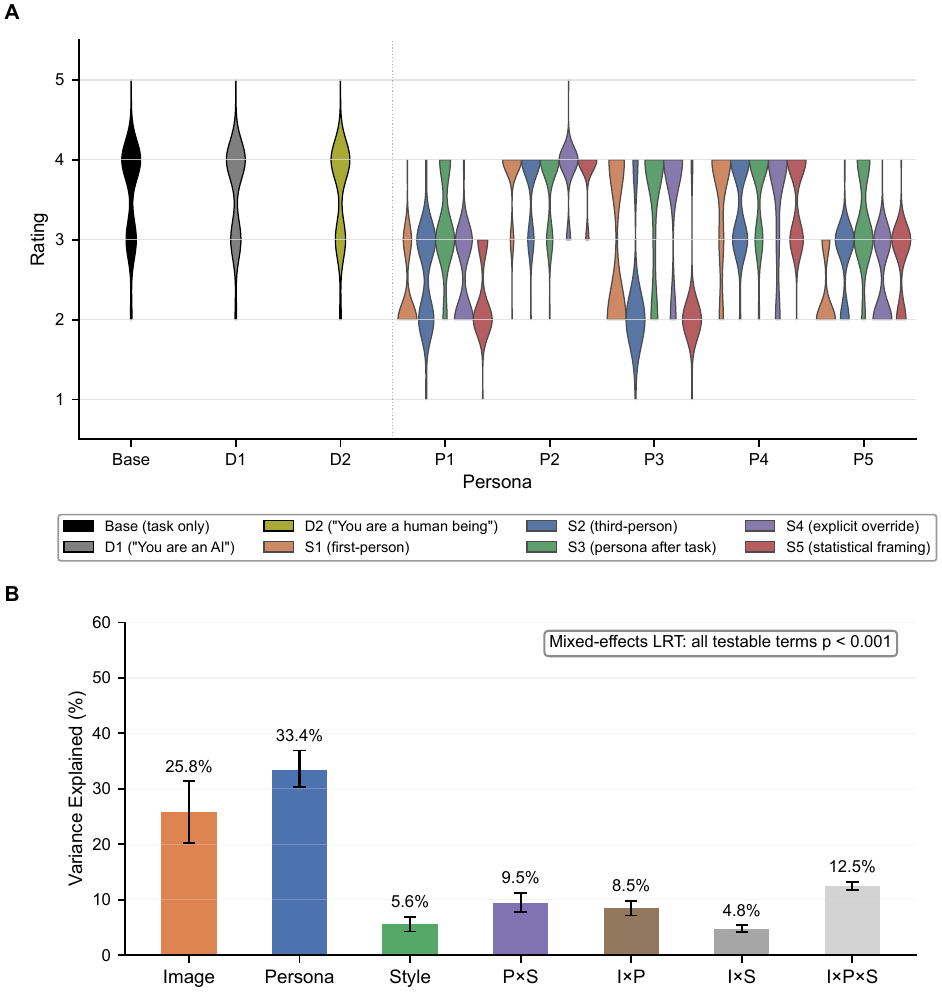}
    \caption{\textbf{Ratings depend more on the persona than on the prompting style (Big Five schema).} (A)~Distributions of the 100 image ratings for the task-only condition (Base), the minimal identity statements ``You are an AI'' (D1) and ``You are a human being'' (D2), and the five Big Five personas (P1--P5); colours indicate the prompting style (S1--S5). The distributions differ more between personas than between the styles that introduce them, and Base, D1 and D2 are indistinguishable. (B) Share of rating variance explained by the three main effects (image, persona, prompting style), their two-way interactions (P$\times$S, I$\times$P, I$\times$S) and the three-way interaction (I$\times$P$\times$S). Bars show the mean over 500 image-level bootstrap replicates, whiskers the 95\% interval. All testable terms were significant in the mixed-effects model ($p < 0.001$; \supptab{supp:tab:variance}). Persona explains the largest share and style the smallest of the three main effects. Most of the persona effect is a shift in a persona's overall rating level, but the I$\times$P term shows that personas also change which images they rate higher. Prompting style, which dominated the global structure of the representations (Fig.~\ref{fig:persona_prompts}), thus has little effect on the ratings, whereas the persona, which structured them only finely, has the largest. The same analysis for the socio-demographic schema is shown in \suppfig{supp:fig:ratings_desc}.}
    \label{fig:ratings}
\end{figure}

\begin{figure}[H]
    \centering
    \includegraphics[width=\linewidth]{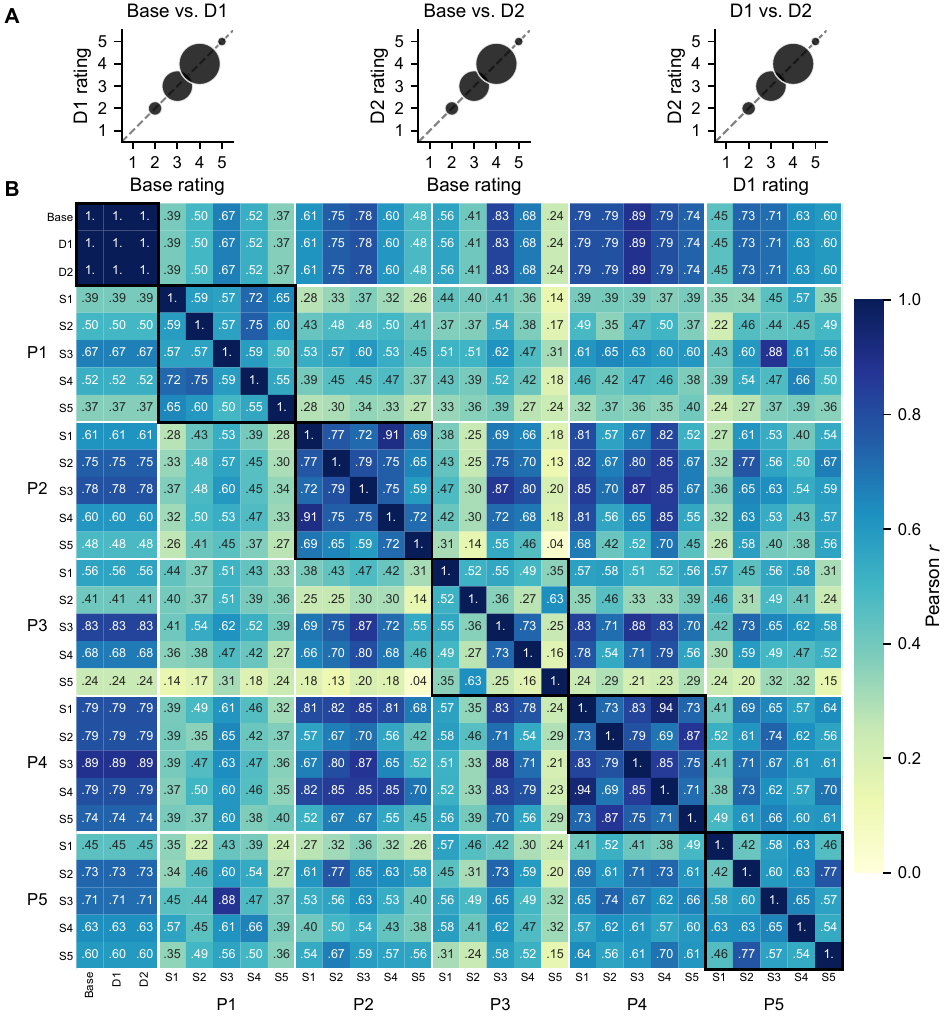}
    \caption{\textbf{Rating patterns follow the persona more than the prompting style, and minimal identity statements leave them unchanged (Big Five schema).} (A)~Ratings under Base and the minimal identity statements D1 and D2 plotted against each other on the 1--5 scale; marker area is proportional to the number of images. The three conditions rate all 100 images identically, although D1 and D2 move the representations away from those of Base (Fig.~\ref{fig:no_prompt}). (B)~Pearson correlations of the 100 image ratings between all 28 conditions: Base, D1, D2 and the 25 persona $\times$ style combinations, ordered by persona (P1--P5) and, within each persona, by style (S1--S5). Each outlined block compares one persona across its five styles. Ratings agree more for the same persona across styles than for the same style across personas, visible as darker blocks along the diagonal that are clearest for P2 and P4; the persona thus shapes which images are rated higher more than the style does. The same analysis for the socio-demographic schema is shown in \suppfig{supp:fig:corr_desc}.}
    \label{fig:ratings_map}
\end{figure}

In the representations, prompting style set the global geometry, whereas the persona contributed a finer structure that emerged only in later layers. In the ratings, this hierarchy was reversed: the rating behaviour was primarily determined by the characteristics of the simulated persona rather than by the way in which the persona was prompted (Figs.~\ref{fig:ratings} and~\ref{fig:ratings_map}).

The distributions of the assigned ratings differed markedly between the Big Five personas (Fig.~\ref{fig:ratings}, A) and, to a lesser degree, between the socio-demographic ones (\suppfig{supp:fig:ratings_desc}, A). The model thus adapted its evaluation of the same images to the character it was instructed to simulate. By comparison, changing the prompting style produced substantially smaller shifts in the rating distributions, despite its pronounced effect on the representations.

This dissociation was confirmed by the variance decomposition. The persona explained 33.4\% of the rating variance, the image 25.8\% and the prompting style only 5.6\% (Fig.~\ref{fig:ratings}, B). Most of the persona effect was a shift in how interesting a persona found the images overall, whereas its effect on which images were rated higher, the image $\times$ persona component, was much smaller. The style mattered mainly by changing how much the personas differed, with the persona $\times$ style component accounting for 9.5\%, more than the style itself. The image $\times$ style component and the three-way interaction accounted for the rest, and all testable components were significant in the mixed-effects model (\supptab{supp:tab:variance}).

Under the socio-demographic schema, the image outweighed the persona, explaining 44.5\% of the rating variance against 16.8\%, while the prompting style, at 5.4\%, remained as small as under the Big Five schema (\suppfig{supp:fig:ratings_desc}, B; \supptab{supp:tab:variance}). The Big Five profiles thus shifted the overall level of the model's judgements more strongly than the socio-demographic profiles.

The minimal identity statements show the dissociation between representations and ratings in its clearest form. Declaring the model an AI (D1) or a human being (D2) moved the representations away from those of the task-only condition (Base). Yet the three conditions gave identical responses to all 100 images, the same rating with word-for-word the same justification. An identity statement without attributes thus changed how the model represented the task but not which images it found interesting (Fig.~\ref{fig:ratings_map}, A).

Correlations between conditions show which images each condition rated higher, independent of its overall rating level (Fig.~\ref{fig:ratings_map}, B). They repeated the ordering found in the variance decomposition. Under the Big Five schema, ratings of the same persona under different styles correlated at 0.63 on average, those of different personas under the same style at only 0.51, barely more than pairs sharing neither (\supptab{supp:tab:agreement}). The socio-demographic schema showed the same ordering. Different formulations of the same persona thus tended to preserve its image preferences.

Overall, the findings show that the size of a representational change does not directly predict its effect on model behaviour. Between persona and prompting style, the ordering reversed. The style reorganised the representations more strongly, yet the persona had the much stronger effect on how the model judged visual interestingness. The minimal identity statements went further and changed the representations without changing a single response.

\needspace{6\baselineskip}
\section{Discussion}
\subsection{Summary}
The present study demonstrates that persona prompting influences a multimodal transformer in a more differentiated way than a simple input-output comparison would suggest. The strongest changes in the model's representations were caused by the way a persona was introduced. By contrast, the psychological and biographical characteristics of the simulated personas produced representational differences several times smaller than those between prompting styles. These differences nevertheless became increasingly visible in later layers, where the representational geometry shifted from being primarily structured by image identity towards increasingly reflecting persona identity, most clearly when the prompt told the model to be the persona. At the level of the output, however, this hierarchy was reversed. Prompting style had only a limited influence on the ratings despite its large effect on the representations. The specific characteristics of the simulated persona had a substantially stronger effect on how the same images were evaluated. Most of this effect consisted of an overall shift in interestingness ratings, but which images were rated higher also followed the persona more than the style. This held for both schemas, although under the socio-demographic schema the image remained the largest single source of rating variance. The minimal identity statements, a persona frame without attributes, mark the limiting case. They shifted the representations but left every response unchanged. Thus, the factors that most strongly reshape the representations are not necessarily those that most strongly determine the model's behaviour. Prompt framing primarily reorganises the geometry of internal representations, whereas persona content more strongly determines behavioural judgements.

\subsection{Interpretation}
Why did large representational changes have little or no effect on the ratings? One possibility is that the model represents more than it uses, as language models have been shown to encode information that does not drive their outputs \cite{elazar2021amnesic, gekhman2025inside}. The shift caused by the minimal identity statements, for instance, was not reflected in the observed responses. It may lie in directions that do not influence the output, or change the output probabilities too little to alter a single greedily decoded token. Post-training could contribute, as it steers models towards a default assistant persona \cite{anthropic2026persona} that corresponds to a direction in activation space \cite{lu2026assistant}. An identity statement may shift the representations without moving the model away from this persona, since it carries nothing that bears on the task, even when it declares the model human. A persona supplies such content, and the ratings follow it.

In-context adaptation may provide a further explanation
for why relatively small representational changes can
nevertheless produce substantial behavioural effects.
Persona information may selectively modulate task-relevant
representations without updating the model's parameters.
Even small changes along behaviourally relevant directions
could alter the resulting judgements, whereas much larger
changes in other directions may leave the responses unchanged.
Thus, behavioural relevance may depend less on the overall
magnitude of representational change than on which dimensions
of the representation space are affected.

\subsection{Comparison with existing literature}
Previous work has shown that personality-related information can be read from hidden activations and steered along dedicated directions \cite{feng2026persona,chen2025persona}, and that induced personality profiles change model responses beyond personality questionnaires \cite{jiang2024personallm, serapio2025psychometric}. Our results extend these findings by showing how image identity, persona content and prompt framing jointly structure the layer-wise representational geometry of a multimodal model, and how this geometry relates to behavioural judgements. They also fit reports that expressed personality and the effects of sociodemographic prompting depend on framing and assigned roles \cite{nagao2026personality, beck2024sensitivity}, except that in our data the framing outweighed the persona in the representations but far less in the ratings.

Personas in the system prompt have been reported not to improve performance on factual questions \cite{zheng2024helpful}. Our minimal identity statements provide a complementary observation: they leave subjective judgements unchanged despite altering the internal representations. Studies relating activations to behaviour have so far examined one kind of factor at a time \cite{chen2025persona, kamruzzaman2026demographic}; comparing persona content with prompt form adds a case in which the factor that dominated the representations was the minor one for the judgements.

That socio-demographic personas shifted the ratings less than Big Five profiles agrees with work on human and simulated judgements. In human annotation data, demographic, social and behavioural variables explain less than 10\% of the variance in subjective judgements \cite{hu2024quantifying, orlikowski2023ecological}. In simulations, demographic agents reproduce individual survey answers less accurately than agents grounded in interviews or personality surveys \cite{park2024llm}, and demographic personas agree closely with one another while missing the views of their groups \cite{yang2026persona}. These findings suggest that psychological profiles may provide stronger constraints on simulated judgements than demographic descriptions, although our two schemas also differ in format.

In vision-language models, persona conditioning is known to change task performance and how images are interpreted \cite{jia2026modeling, da2026analyzing}. We extend these findings by tracing the layer-wise emergence of persona-related representational geometry and demonstrating that the magnitude of representational change does not necessarily reflect its behavioural relevance.

\subsection{Limitations and Future Directions}\label{sec:limitations}
Our study measures persona \emph{sensitivity}, not persona \emph{fidelity}. The dissociation between representations and ratings concerns the model itself, but whether the simulated perspectives resemble those of the people they describe requires ratings from matching human viewers. This matters because \gls{llm} simulations are known to flatten identity groups \cite{wang2025large}, to drift towards caricature \cite{cheng2023compost} and to reflect some populations better than others \cite{santurkar2023whose}. The percentile profiles were meant to discourage caricature, but their larger effect could equally reflect exaggeration.

All results come from a single 8B model with greedy decoding, and larger models, other architectures or sampled responses may weigh persona and framing differently. Each judgement was a single ordinal rating, a constrained format in which persona effects may be stronger than in open-ended responses, as reported for activation steering \cite{frising2025linear}. Comparing models before and after post-training could clarify the role of the default assistant persona, while causal interventions could identify which representational dimensions actually influence behavioural judgements. In particular, targeted perturbations could test whether small changes along task-relevant representational dimensions produce stronger behavioural effects than larger changes in functionally irrelevant directions. Systematically varying the amount and task relevance of persona information could further reveal whether in-context adaptation selectively engages behaviourally relevant dimensions of the representation space.

Each schema contains five prototypical personas, enough to separate persona from framing but not to map individual traits or demographic variables, which a systematic variation of single attributes could do. Each prompting style had a single wording, and the styles are not orthogonal, since S3 changes both the position of the persona and its introduction and S5 removes the role-play while adding a statistical claim. Part of the differences between styles may therefore reflect the particular prompts, which several wordings per style and a separate manipulation of position would resolve. The two schemas, finally, differ in vocabulary (verbal versus numeric) and content (biography versus traits), which only a crossed design could separate. Beyond these limitations, persona-conditioned models could support personalised predictions of interestingness, while ensembles of personas could complement the task-only baseline by providing a reference that captures variation across simulated perspectives.

\subsection{Concluding Remarks}
The central finding of this study is a dissociation between representational change and behavioural relevance. A persona's frame reorganised the representations of a multimodal model, and even a bare statement that the model is an AI or a human shifted them, yet the model's judgements of the images followed what the persona contained. Two practical consequences follow. Studies that simulate people with personas should report and vary the prompt form, because it changes how strongly a persona shows in the judgements. Studies that read personas from activations should check representational differences against behaviour, because the most prominent ones need not be the ones the model uses. 
More broadly, consistent with the simulator view of language models \cite{janus2022simulators, anthropic2026persona}, these findings suggest that a model may represent multiple context-dependent perspectives without all of them influencing its behaviour. Distinguishing such latent representations from behaviourally effective ones may be important not only for persona simulation but also for understanding implicitly emerging personas and their implications for alignment \cite{betley2025emergent, betley2025weird}.

\clearpage

\backmatter

\bmhead{Data accessibility}
\stmtdata

\bmhead{Declaration of AI use}
\stmtai

\bmhead{Funding}
\stmtfunding

\bmhead{Acknowledgements}
\stmtack

\begin{appendices}

\section{Derivation of Psychometric Profiles}
\label{supp:sec:psycho}
The psychometric profiles of the Big Five schema are stylised percentile vectors designed to provide the model with distinct, theoretically grounded trait configurations. Rather than sampling traits at random, these profiles were constructed to reflect qualitative patterns identified in established psychometric research. To facilitate computational modelling, the raw empirical values (which are typically reported as continuous Z-scores) were translated into idealised percentiles rounded to the nearest multiple of five.

In the five-factor model \cite{mccrae1992introduction}, openness captures curiosity and receptivity to novel experiences, conscientiousness organisation and self-control, extraversion sociability and energy, agreeableness compassion and cooperativeness, and neuroticism the tendency towards negative affect and stress.

\paragraph{Profiles derived from large-scale data clustering.}
Three of the profiles map onto the robust personality types identified by Gerlach et al. \cite{gerlach2018robust}, who analysed multi-million-user datasets to find replicable trait clusters across populations.

\begin{description}
  \item[The Reserved (P1)] Mirrors the ``Reserved'' cluster, characterised by emotional stability (low Neuroticism) and low Openness and Extraversion. Conscientiousness and Agreeableness, only moderately elevated in that cluster, are set high here.
  \item[The Role Model (P2)] Captures the ``Role Model'' cluster, which exhibits socially desirable traits across all dimensions. This profile features high levels of Extraversion, Openness, Agreeableness, and Conscientiousness, alongside very low Neuroticism.
  \item[The Normative (P5)] Corresponds to the ``Average'' cluster in the Gerlach typology. It reflects a highly typical population configuration with moderately high Neuroticism and slightly lower Openness. Extraversion, high in that cluster, is kept average here.
\end{description}

\paragraph{Profiles derived from additional frameworks.}
To ensure a comprehensive coverage of the Big Five trait space, two additional profiles were synthesised from distinct psychological traditions.

\begin{description}
  \item[The Undercontrolled (P3)] Rooted in the person-centred developmental typologies of ego-control and ego-resiliency \cite{robins1996resilient}. The undercontrolled phenotype is typically characterised by low Conscientiousness and low Agreeableness paired with high Extraversion and Neuroticism, reflecting a tendency towards impulsivity and externalising behaviours.
  \item[The Highly Sensitive (P4)] Drawn from the framework of Sensory Processing Sensitivity \cite{aron1997sensory}. This profile models individuals who process internal and external stimuli deeply. It is defined by high Openness (reflecting aesthetic sensitivity and intellectual curiosity), high Agreeableness (strong empathy), and high Neuroticism (susceptibility to overstimulation and stress).
\end{description}

By grounding the synthetic archetypes in these diverse taxonomies, the experimental manipulation covers a realistic yet systematically dispersed range of human behavioural tendencies without relying on unprincipled trait combinations.

\section{Socio-demographic Personas}
\label{supp:sec:descriptive}

\begin{figure}
    \centering
    \includegraphics[width=\linewidth]{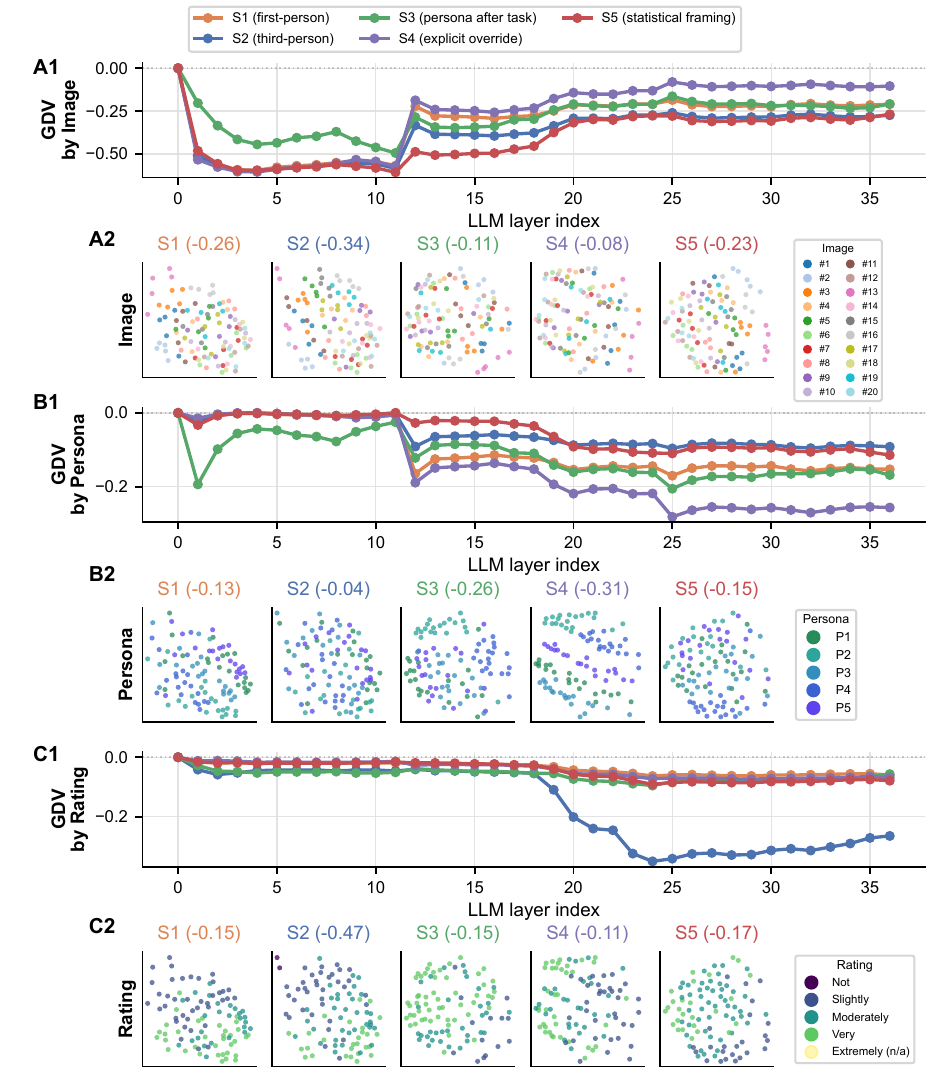}
    \caption{\textbf{Layer-wise transition from image to persona and rating structure in Qwen3-VL-8B (socio-demographic schema).} As in \mainfig{fig:GDV_prompts}, \gls{gdv} values quantify clustering in the original high-dimensional hidden-state space at every decoder layer; more negative values indicate stronger cluster separation, and each curve is one prompting style (S1--S5). The projection panels show two-dimensional MDS projections at layer~36 for a 20-image subset, one per style, with the GDV of that projection in each title. (A1)~GDV by image. Image clustering dominates the early layers and weakens abruptly at layer~12 under every style except S5, where it weakens only around layer~20. (A2)~Projections coloured by image. (B1)~GDV by persona (P1--P5). Persona clustering emerges at layer~12, earlier than for the Big Five personas, abruptly under every style except S5, where it grows slowly. It is strongest under S4, apart from a peak under S3 at layer~1 that falls off sharply at layer~2. (B2)~Projections coloured by persona. (C1)~GDV by assigned rating. Rating clustering increases markedly in later layers only under S2. (C2)~Projections coloured by rating; \emph{Extremely Interesting} is marked absent because it does not occur in the 20-image subset.}
    \label{supp:fig:gdv_desc}
\end{figure}

\begin{figure}
    \centering
    \includegraphics[width=\linewidth]{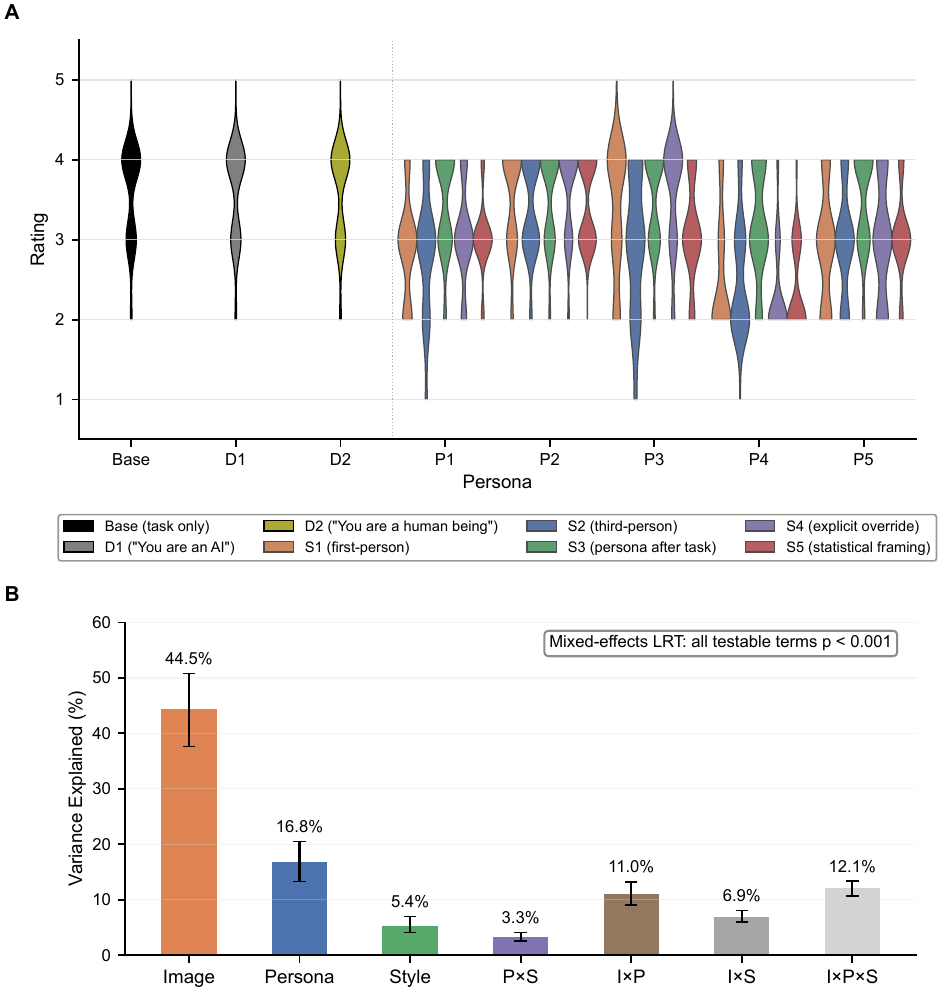}
    \caption{\textbf{Ratings depend most on the image, then on the persona, and least on the prompting style (socio-demographic schema).} (A)~Distributions of the 100 image ratings for Base, the minimal identity statements D1 and D2, and the five socio-demographic personas (P1--P5); colours indicate the prompting style (S1--S5), as in \mainfig{fig:ratings}. The distributions differ less between personas than under the Big Five schema, with P4 rating lowest. (B)~Share of rating variance by component, as in \mainfig[B]{fig:ratings}; all testable terms were significant in the mixed-effects model ($p < 0.001$; Table~\ref{supp:tab:variance}). Unlike under the Big Five schema, the image explains the largest share and the persona the second largest, while style remains the smallest main effect; personas change which images they rate higher (I$\times$P) slightly more than under the Big Five schema.}
    \label{supp:fig:ratings_desc}
\end{figure}

\begin{figure}
    \centering
    \includegraphics[width=\linewidth]{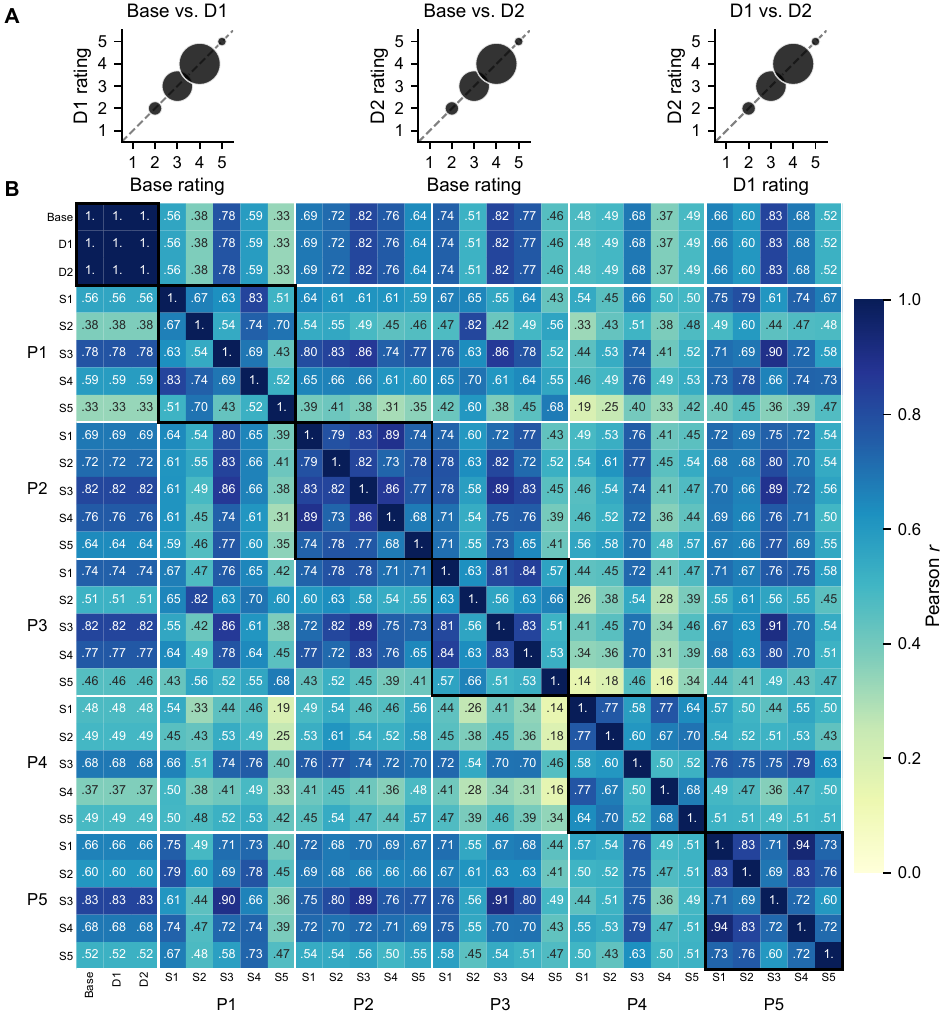}
    \caption{\textbf{Rating patterns follow the persona more than the prompting style (socio-demographic schema).} (A)~Ratings under Base and the minimal identity statements D1 and D2 plotted against each other, as in \mainfig[A]{fig:ratings_map}. (B)~Pearson correlations of the 100 image ratings between all 28 conditions, ordered as in \mainfig[B]{fig:ratings_map}. As under the Big Five schema, ratings agree more for the same persona across styles than for the same style across personas, with the clearest blocks for P2 and P5, and they stay closest to those of Base when the persona is placed after the task (S3).}
    \label{supp:fig:corr_desc}
\end{figure}

The main text reports the Big Five schema, whose numeric trait profiles make the persona manipulation explicit and quantitative. Every analysis was run identically on the socio-demographic schema, whose personas are specified by country of origin, age band, gender, professional background, and transient state. This \suppunit\ reports that replication: the layer-wise clustering measures (Fig.~\ref{supp:fig:gdv_desc}), the rating distributions and variance decomposition (Fig.~\ref{supp:fig:ratings_desc}), and the cross-condition agreement structure (Fig.~\ref{supp:fig:corr_desc}).

The qualitative pattern of the main text holds under the socio-demographic schema: prompt style dominates the global organisation of the representational space while persona identity emerges in the later decoder layers. The two schemas differ in one respect that deserves attention. Under the Big Five schema the persona accounts for the largest share of rating variance, whereas under the socio-demographic schema image content does, with the persona second (Table~\ref{supp:tab:variance}). A socio-demographic profile therefore shifts the model's judgements less forcefully than an explicit trait vector, which is consistent with the Big Five schema stating the psychological disposition directly rather than leaving it to be inferred from biography.

\section{Clustering by Prompting Style}
\label{supp:sec:style_gdv}
\suppfigreset

\begin{figure}
    \centering
    \includegraphics[width=\linewidth]{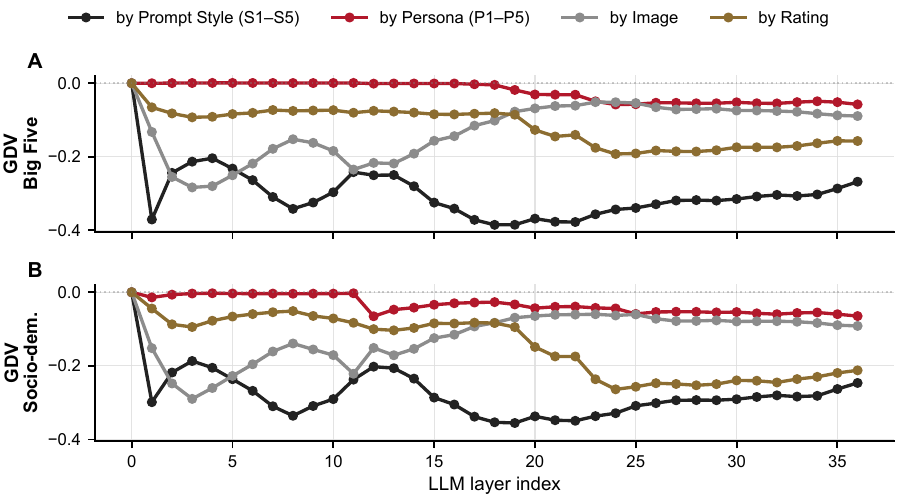}
    \caption{\textbf{Clustering by prompting style exceeds clustering by persona at every layer.} \gls{gdv} in the original hidden-state space at every decoder layer, computed on the representations of all five prompting styles together (2{,}500 per layer) and grouped by prompting style (S1--S5), by persona (P1--P5), by image and by assigned rating; more negative values indicate stronger clustering. (A)~Big Five schema. (B)~Socio-demographic schema. Style and persona each form five classes of equal size, so their values are directly comparable. Clustering by style is strong from the first layers on and peaks around layers 18--22, whereas clustering by persona appears only in later layers and remains several times weaker. Rating clustering reflects persona and style together, since both shift the rating level.}
    \label{supp:fig:style_gdv}
\end{figure}

The per-style curves in \mainfig{fig:GDV_prompts} and Fig.~\ref{supp:fig:gdv_desc} each contain a single prompting style and therefore cannot compare clustering by style with clustering by persona. Fig.~\ref{supp:fig:style_gdv} computes the \gls{gdv} on the representations of all five styles together. Because the styles separate the representations, each persona and each image now falls into five style-specific sub-clusters, so their \gls{gdv} values are smaller in magnitude than in the per-style curves. The values in Fig.~\ref{supp:fig:style_gdv} should therefore be compared with each other rather than with those in \mainfig{fig:GDV_prompts} and Fig.~\ref{supp:fig:gdv_desc}, whereas the layers at which persona clustering emerges and image clustering recedes are the same.

\section{Variance Decomposition, Model Comparison and Rating Agreement}
\label{supp:sec:numbers}

Table~\ref{supp:tab:variance} tabulates the quantities plotted in \mainfig[B]{fig:ratings} and Fig.~\ref{supp:fig:ratings_desc}B for both schemas, so that the numbers can be read off without measuring bars. The two schemas are decomposed separately (\mainsec{sec:variance_decomposition}), each over $100 \times 5 \times 5 = 2{,}500$ ratings.

Two contrasts are worth naming. First, the persona share roughly doubles from the socio-demographic to the Big Five schema (16.8\% to 33.4\%) and overtakes image content, while the prompt-style share is essentially identical in the two schemas (5.4\% versus 5.6\%): what the persona is stated to be matters far more than how it is introduced, and this holds for both schemas. Second, the persona~$\times$~style interaction is roughly three times larger under the Big Five schema, meaning trait profiles are not merely stronger but also more sensitive to framing than biographical ones.

Every term is significant by likelihood-ratio test in both schemas, including the interaction, so the small percentages attached to the style terms should be read as small effects that are reliably present rather than as absent effects.

Table~\ref{supp:tab:agreement} summarises the correlation matrices of \mainfig[B]{fig:ratings_map} and Fig.~\ref{supp:fig:corr_desc}B by what a pair of conditions shares, for both schemas.

\begin{table*}[t]
  \centering
  \small
  \caption{Rating variance and mixed-effects model comparison, per schema. \textbf{(a)}~Sum-of-squares variance decomposition over the crossed persona $\times$ prompt-style cells; entries are the mean over 500 image-level cluster-bootstrap replicates with the 2.5th--97.5th percentile interval, i.e. exactly what \mainfig[B]{fig:ratings} and Fig.~\ref{supp:fig:ratings_desc}B plot. With one rating per image, persona and style, the three-way interaction is the residual; it is not sampling noise, because decoding is greedy. \textbf{(b)}~Likelihood-ratio tests of the mixed-effects model (\mainsec{sec:mixed_effects}) against the nested model omitting each term.}
  \label{supp:tab:variance}
  \textbf{(a) Variance explained (\%), mean [95\% CI]}\\[2pt]
  \begin{tabular}{@{}lcc@{}}
    \toprule
    \textbf{Component} & \textbf{Socio-demographic} & \textbf{Big Five} \\
    \midrule
    Image                        & 44.5 [37.6, 50.8] & 25.8 [20.2, 31.4] \\
    Persona                      & 16.8 [13.3, 20.5] & 33.4 [30.3, 36.9] \\
    Prompt style                 &  5.4 [\ 4.1, \ 7.0] &  5.6 [\ 4.3, \ 7.0] \\
    Persona $\times$ style       &  3.3 [\ 2.6, \ 4.1] &  9.5 [\ 7.8, 11.3] \\
    Image $\times$ persona       & 11.0 [\ 9.1, 13.2] &  8.5 [\ 7.2, \ 9.8] \\
    Image $\times$ style         &  6.9 [\ 6.0, \ 8.1] &  4.8 [\ 4.2, \ 5.4] \\
    Image $\times$ persona $\times$ style & 12.1 [10.7, 13.3] & 12.5 [11.7, 13.2] \\
    \bottomrule
  \end{tabular}

  \vspace{0.9em}
  \textbf{(b) Likelihood-ratio tests}\\[2pt]
  \begin{tabular}{@{}lcccc@{}}
    \toprule
    & \multicolumn{2}{c}{\textbf{Socio-demographic}} & \multicolumn{2}{c}{\textbf{Big Five}} \\
    \cmidrule(lr){2-3}\cmidrule(lr){4-5}
    \textbf{Term} & $\chi^2$ (df) & $p$ & $\chi^2$ (df) & $p$ \\
    \midrule
    Persona                & 367.1 (4)  & $<0.001$ & 633.3 (4)  & $<0.001$ \\
    Prompt style           & 224.2 (4)  & $<0.001$ & 303.7 (4)  & $<0.001$ \\
    Persona $\times$ style & 370.8 (16) & $<0.001$ & 889.7 (16) & $<0.001$ \\
    Image $\times$ persona (variance) & 331.0 (1) & $<0.001$ & 192.3 (1) & $<0.001$ \\
    Image $\times$ style (variance)   & 128.1 (1) & $<0.001$ &  31.5 (1) & $<0.001$ \\
    \bottomrule
  \end{tabular}
\end{table*}

\begin{table}[t]
  \centering
  \small
  \caption{Mean pairwise Pearson correlation of the 100 image ratings, per schema, averaged over the off-diagonal cells of \mainfig[B]{fig:ratings_map} and Fig.~\ref{supp:fig:corr_desc}B. The rows group the 25 persona $\times$ style conditions by what a pair shares.}
  \label{supp:tab:agreement}
  \begin{tabular}{@{}lcc@{}}
    \toprule
    \textbf{Pairs of conditions} & \textbf{Socio-demographic} & \textbf{Big Five} \\
    \midrule
    Same persona, different style   & .69 & .63 \\
    Same style, different persona   & .62 & .51 \\
    Neither shared                  & .56 & .48 \\
    \bottomrule
  \end{tabular}
\end{table}

\clearpage

\end{appendices}

\bibliography{sn-bibliography}%

\end{document}